\documentclass[manuscript,screen]{acmart}
\AtBeginDocument{%
  }

\setcopyright{acmlicensed}
\copyrightyear{2026}
\acmYear{2026}
\acmDOI{XXXXXXX.XXXXXXX}

\begin{document}

\title{AI as Entertainment}

\author{Cody Kommers}
\authornote{Corresponding Author}
\email{ckommers@turing.ac.uk}
\orcid{0009-0007-8985-0085}
\affiliation{%
  \institution{The Alan Turing Institute}
  \city{London}
  \country{UK}
}

\author{Ari Holtzman}
\affiliation{%
  \institution{University of Chicago}
  \country{USA}}



\begin{abstract}
Generative AI systems are predominantly designed, evaluated, and marketed as intelligent systems which will benefit society by augmenting or automating human cognitive labor, promising to increase personal, corporate, and macroeconomic productivity. But this mainstream narrative about what AI is and what it can do is in tension with another emerging use case: entertainment. We argue that the field of AI is unprepared to measure or respond to how the proliferation of entertaining AI-generated content will impact society. Emerging data suggest AI is already widely adopted for entertainment purposes---especially by young people---and represents a large potential source of revenue. We contend that entertainment will become a primary business model for major AI corporations seeking returns on massive infrastructure investments; this will exert a powerful influence on the technology these companies produce in the coming years. Examining current evaluation practices, we identify a critical asymmetry: while AI assessments rigorously measure both benefits and harms of intelligence, they focus almost exclusively on cultural harms. We lack frameworks for articulating how cultural outputs might be actively beneficial. Drawing on insights from the humanities, we propose ``thick entertainment'' as a framework for evaluating AI-generated cultural content---one that considers entertainment's role in meaning-making, identity formation, and social connection rather than simply minimizing harm. We argue that this kind of positive vision for AI as entertainment is missing from the field's mainstream discourse, which tends to assume that the main impact of this technology will be based on its capacity for intelligent behavior. We consider an alternative perspective, in which the main impact will come from its capacity to divert, amuse, tell stories, or just help pass the time. While AI is often touted for its potential to revolutionize productivity, in the long run we may find that AI turns out to be as much about ``intelligence'' as social media is about social connection. 
\end{abstract}

\ccsdesc[500]{Applied computing~Media arts}
\ccsdesc[100]{Computing methodologies~Philosophical/theoretical foundations of artificial intelligence}
\ccsdesc[100]{Human-centered computing~HCI theory, concepts and models}

\keywords{Generative AI, Entertainment, Culture, LLMs, Societal Impact, Meaning-making}


\maketitle

\section{Introduction}

A foundational assumption in the way Generative AI (GenAI) systems are designed, evaluated, and marketed is that their main attribute is intelligence. For example, this is reflected in mainstream approaches to AI benchmarking: a system’s outputs are graded in much the same way as a student exam, with right and wrong answers summarized in a percentage score denoting how close they come to some idealized response \cite{guo2023evaluating, burden2025paradigms}. These systems are often marketed as useful in any circumstance, with their capabilities (present or future) sold as like what humans have---only more powerful. 

But just because we call these systems by the name artificial \textit{intelligence} does not mean their defining feature is a kind of all-purpose instrumental utility. By comparison, social media platforms initially began with the intention of connecting people but have steadily been getting less social over time \cite{ryan2017social}. People spend increasingly less time connecting with friends and more following celebrities or passively consuming content  \cite{kavanagh2019more}. This has led some commentators to suggest they would be more accurately called \textit{algorithmic} media: that the main dimension in which these platforms have impacted society is not social connection, but the algorithmic manipulation of attention.\footnote{https://www.nytimes.com/2025/12/07/opinion/meta-facebook-ruling-algorithms.html} Similarly, phones are no longer defined mainly by their faculties of telecommunication. The primary use of a technology is not determined by the name we give it.

Another way of conceptualizing what GenAI systems do is not as intelligence, but as culture \cite{kommers2025computational, farrell2025large, klein2025provocations}. The performance of these systems no longer cleaves to clearly articulated tasks with well-defined value functions. Instead, they are open-ended systems designed to produce a variety of outputs, often across multiple modalities, in an unbounded space of possible social, cultural, or economic contexts. By comparison, humans do not just use their cognitive faculties for productivity in their jobs; they also use their intelligence to tell stories or jokes, bond with friends, denounce enemies, express themselves with fashion, compose moving or enlivening music, create compelling visuals, and make and share things for no other reason than they find it amusing to do so. GenAI can now participate, often quite centrally, in these kinds of cultural processes. But this fact is easily overlooked in a narrative about AI that focuses mainly on how it will supercharge productivity at work.

We present an alternative narrative: AI as entertainment. GenAI is not just a proxy or supplement for human goal-directed action and problem solving. It is a technology whose ability to flexibly generate multimedia content will transform the way we create, share, and understand culture \cite{brinkmann2023machine}. We use the term ``entertainment'' (rather than ``culture'') to reflect the broader economic system of production, the observation that much entertaining content is not lauded for its cultural merit, and the sense that it is often seen not as a primary activity in its own right, but something used to pass the time between the goal-directed tasks from which people typically earn a living. 

Not all entertaining content is created equal. Entertainment can inspire everything from innocuous amusement to clamoring rage, from profound feelings of empathy or kinship to challenging reflections on one's own worldview. For example, social media did not upend important societal trends by selling enterprise software. It did so by subtly modifying the incentives for the creation and distribution of entertaining content in YouTube videos, kitschy memes, and a barrage of attention grabbing text \cite{fisher2022chaos}. At a societal level, it matters what kind of entertainment is incentivized and made possible by our most powerful sociotechnical systems. As AI is incorporated into cultural production across different mediums---in film, news, books, music, the internet, and beyond---it will influence the way we tell and interpret stories, in turn shaping our collective processes for meaning-making, identity, and social connection. 

And yet, we have little idea of what it would mean to succeed---in a technical or societal sense---in developing AI as entertainment in any beneficial way. The crucial difference between these narratives (AI ``doing'' intelligence for productivity versus culture for entertainment) is that the field of AI is built around the development and evaluation of intelligence. There is much disagreement about the conditions under which these systems can be used as effective thought partners \cite{collins2024building} or how intelligent they might really be \cite{lake2017building}. But there is nonetheless a shared language for articulating and debating AI’s potential to impact society by automating or augmenting human cognitive labor. The same is not true for culture. There is a thriving discourse about what AI can get wrong: with discussions of safety, ethics, fairness, morality, transparency, and governability \cite{lazar2023ai, weidinger2022taxonomy, anwar2024foundational, kasirzadeh2025characterizing}. But we do not yet have a shared language for a positive vision of cultural AI: how could AI potentially influence the generation or consumption of culture in a way that represents an active improvement for society, rather than a mere absence of harm?

Our position is not that AI \textit{should} be used for entertainment; it is that AI already is, and increasingly will be, used for entertainment---and that the field is largely unprepared to measure, shape, or respond to the massive societal consequences that will come from this. In this paper, we consider evidence that people, especially kids, already use AI for entertainment. We also argue that entertainment will be a significant business model for big AI corporations. We contend that this consideration, more than any other, will impact the technologies produced by them in the coming years. 

This paper is meant to give an initial vision of what it would look like to prepare to evaluate the cultural, political, and psychological dimensions of AI as entertainment. Thinking about AI solely in terms of instrument utility risks overlooking the fact that entertainment is a crucial site of meaning-making \cite{kommers_sense-making_2025,Bruner1990}, and the role that AI will have in shaping the narratives and images that shape our collective sense of identity, morality, and purpose \cite{bisk2020experience}. What that role is and how it affects society is not yet determined---but the window to intentionally design these consequences is rapidly closing. We would rather prepare for them in advance than study them in retrospect.

\section{People already use AI for entertainment}

Readers of this paper might be surprised about the extent to which AI is already being used for entertainment. Most academics probably do not find AI-generated content especially compelling. Whether on the internet or at work, overtly AI-generated content is often derided as AI slop \cite{kommers2025slopmatters}. Much of the academic community is skeptical about the rapid adoption of AI, with entertainment coming in as a low priority on the list of capabilities we would like to cultivate in leading AI systems. Furthermore, entertainment itself is not a theme that typically enjoys widespread academic interest outside areas such as media studies \cite{postman1985amusing}. When this space of topics is approached, it is usually from the more academic lens of ``culture.'' All this leads to the assumption that entertainment is not a primary use case for AI---and even if it is, it is not especially worth studying. But this position becomes less tenable in the face of mounting evidence that the ecosystem of contemporary AI is increasingly coming to be architected around entertainment.

One example that will be familiar to many readers is Character.AI. This is a platform that offers users an interface to converse with anyone they would like to talk to---from pop culture figures, to fictional characters, to historical sages---via an AI-generated chatbot attempting to assume their characteristics or perspective. Some of these characters will be broadly familiar: such as the Forumla 1 driver Lewis Hamilton, Harry Potter’s nemesis Draco Malfoy, or world-historically gifted interlocutor Socrates. Others are more specific to the cultural practices of the platform, often a mix of internet culture and Japanese anime aesthetics. For example, popular accounts include Guy Best Friend, whose stated attributes are ``overprotective, sarcastic, funny, deep voice, TALL'' (95 million interactions); Stella: a catty AI ``assistant'' who responds with eye-rolling snark whenever you ask her to do something (73 million); AwkwardFamilyDinner: ``the chaos is real, and so is the tension'' (20 million); and Pokemon RPG: ``Be human, or even a Pokémon!'' (10.5 million). A report from August 2025 estimated that Character.AI has 20-28 million monthly active users, about 53\% of whom are under the age of 25 \cite{Lee2025}.

A less well known example might be the emergence of virtual YouTubers, also known as Vtubers \cite{ye2025my}. These are YouTube accounts whose content is solely generated by AI. They typically feature an avatar---again, frequently inspired by aesthetics from anime---who discourses about the kind of things that human YouTubers discourse about (the meaning of life, what they had for breakfast, etc). A representative example is Neuro-sama, streaming from the Twitch channel vedal987, whose videos are archived on YouTube and routinely rack up millions of views. Neuro-sama’s ``life'' resembles a Sailor Moon-style protagonist living as an otherwise normal teenage girl who uses the internet. Some videos feature her ``Trying Weird Foods from Trade Joe’s'' while others feature her in music videos. Viewers actively participate in the creation of content by paying to influence which activities Neuro-sama engages in next \cite{ye2025my}. Recently, Neuro-sama enjoyed a stint as the most popular channel on the live streaming platform Twitch, with 165,268 actively paying subscribers.\footnote{https://futurism.com/artificial-intelligence/ai-twitch-streamer-neuro-sama}

These are two of the many platforms in which AI is being used for entertainment---but they are far from the only ones. To take just a few others: 
\begin{itemize}
  \item In short-form video, OpenAI recently released Sora, an AI-first TikTok-style platform on which users can generate and share AI-generated videos. 
  \item In music, AI ``artists'' have begun to break into mainstream listening trends, with an AI-generated country song topping the US Billboard charts for the first time via streams on Spotify.\footnote{https://www.theguardian.com/technology/2025/nov/13/ai-music-spotify-billboard-charts} 
  \item In long-form spoken audio, the most popular feature of Google’s Notebook LM, and arguably the one that most sets it apart from competitors, is a function for turning content into an AI-generated podcast.\footnote{https://www.wsj.com/tech/ai/google-notebooklm-ai-podcast-deep-dive-audio-c30a06b3} 
  \item In fiction, it is increasingly unclear whether the median reader still prefers fictional stories generated by humans over those generated by AI.\footnote{https://www.newyorker.com/culture/the-weekend-essay/what-if-readers-like-ai-generated-fiction}
\end{itemize}

It is perhaps unsurprising---though no less distressing---to note that children and adolescents are disproportionately likely to use AI for entertainment. In a survey of 1,060 teens (age 13-17) in the United States from Spring 2025 \cite{robb2025talk}, 73\% reported having used AI companions (such as Character.AI, Replika, or other chatbots designed for parasocial diversion rather than productivity). Just over half of the teens surveyed claimed to be regular users, engaging with these chatbots at least a few times per month. When asked why they use these chatbots, the most common response was because ``It's entertaining'' (30\%). 

This effect goes beyond systems designed for the specific use case of companionship. Another survey from the same period looked at the use of chatbots (e.g., ChatGPT, Gemini, or Snapchat’s My AI) among 1,000 UK children and teens, age 9-17 \cite{bunting2025myself}. A full quarter of the respondents said they used these systems ``Just for fun / escapism (e.g. playing games, roleplaying)''. Despite the terms and conditions of these platforms requiring users to be at least 13 years old, 58\% of surveyed children age 9-12 reported having used an AI chatbot. The third most common system used by these children (after ChatGPT and Gemini) was Snapchat's My AI---which, presumably, is less likely to be used for information seeking (e.g., doing homework) or on a parent's account. 

But the proliferation of AI as entertainment is not limited solely to teens and adolescents. Emerging data suggests people may be more motivated to use AI for entertainment than for productivity. In a study that looked at how participants used AI over five weeks \cite{chandra2025longitudinal}, active users were given a daily prompt to use a specific AI system and subsequently filled out a ``Motivation for AI Use'' survey \cite{huang2024ai}. One subcomponent of the survey measured motivation to use AI for instrumental utility, another for entertainment. In the sample of active users, self-reported motivation to use AI for instrumental utility leveled off after the first week, with an increase of approximately 20\%. By contrast, self-reported motivation to use AI continued to increase for a longer period and reached a higher overall magnitude, leveling off around 30\%. 

More data are needed to provide a fuller picture, but this suggests that people may find it easier---and, potentially, less disappointing---to integrate AI into their routine for entertainment than productivity. At face value, the promise of easing one's workload with AI is alluring. But this high potential payoff can also lead to disappointment when an AI fails to live up to expectations. The constraints of AI as entertainment are more fungible. This is a case where AI's difficult-to-eradicate hallucination problem can be a value-add, rather than a liability, when its output is not committed to accurately representing some ground truth aspect of reality. To this end, a report from the UK's AI Security Institute found that 33\% of the 2,028 UK residents sampled had used AI for some non-instrumental purpose---such as companionship, emotional support, or social interaction---in the past year.\footnote{https://www.aisi.gov.uk/frontier-ai-trends-report \textit{see page 39}} It is an industry truism that today’s AI is the worst it will ever be; it is also the least entertaining it will ever be.

\section{Entertainment as a business model for AI}

Academics tend not to take a position on how a given capability will affect the bottom line of AI companies. Typically, they try to influence the societal impact of technology by regulation \cite{anderljung2023frontier, simbeck2022facct}. While thoughtful, robust regulation is crucial for the beneficial deployment of AI, it is only one means of shaping technology. Even with thoughtful regulation in place, big corporations are highly motivated to circumvent the limitations imposed by such regulatory strictures and put a large amount of resources toward doing so. But the stick is only one way to influence technological development; the carrot is another. The single largest decision that will affect the future development of AI is how the companies producing it plan to generate the massive return on investment needed to justify the amount of money expended on developing their products. 

The amount of money which has recently been invested in GenAI is, to put it mildly, a lot. Recent reporting from the \textit{New York Times} shows that large tech infrastructure companies like Nvidia and Oracle have agreed to provide 20 gigawatts of computing power to OpenAI over the next decade.\footnote{https://www.nytimes.com/2025/10/07/business/dealbook/openai-nvidia-amd-investments-circular.html} This amount of power is equivalent to about 20 nuclear reactors; the estimated cost of providing it is in the ballpark of \textdollar1 trillion. For context, at the time these deals were struck, OpenAI was valued at \textdollar500 billion, while having only made \textdollar1.8 billion in profit during the first half of 2025. And that's just one company. Global corporate investment in AI is estimated to have reached \textdollar252.3 billion in 2024 alone \cite{maslej2025artificial}. Likewise, a McKinsey survey found that 88\% of respondents reported GenAI was being used in their business.\footnote{https://www.mckinsey.com/capabilities/quantumblack/our-insights/the-state-of-ai} The market capitalization of Nvidia---a kind of barometer for the AI industry as a whole---is more than \textdollar4.5 trillion, reflecting an increase of more than 1,350\% from five years ago.\footnote{https://uk.finance.yahoo.com/quote/NVDA/history/} 

From this lens, the enemy of fairness, accountability, and transparency is a business model in which these considerations are illegible---or, worse, antithetical. Academics should take an active interest in exploring how AI can make money in a way that best aligns with societally beneficial considerations. As a point of comparison, plant-based meat alternatives might not exist if people did not seek to fulfill the relevant market demand while also optimizing for animal welfare and sustainable production. The alternative is to leave a vacuum: to let purely corporate interests determine the path to profitability, likely with minimal incentive to consider how downstream effects might be detrimental to society beyond the time horizon of a quarterly earnings report. 

It is unclear whether the strategy of developing AI tools for workplace productivity will be sufficient to generate the required level of revenue. If AI revolutionizes a wide range of industries and obviates the need for human cognitive labor across swathes of the economy, then the corporations whose technologies power this economic revolution will probably be profitable. But a revolution of this magnitude is still hypothetical. If these gains do not materialize, or if the profitability is less than expected, alternative revenue streams will be needed.\footnote{https://www.noahpinion.blog/p/americas-future-could-hinge-on-whether}

Taken together, entertainment-based industries offer markets of the size needed to earn back the money invested in AI. Rough estimates for the annualized revenue in relevant industries include: music streaming (\textdollar20 billion\footnote{https://www.ifpi.org/wp-content/uploads/2024/03/GMR2025\_SOTI.pdf}), news (\textdollar85 billion\footnote{https://www.kenresearch.com/global-newspaper-market}), publishing (\textdollar90 billion\footnote{https://online.hull.ac.uk/blog/top-trends-transforming-the-future-of-publishing}), long-form video (e.g., Netflix; \textdollar115 billion\footnote{https://finance.yahoo.com/news/live-streaming-market-size-surpass-073900911.html}), short-form video (e.g., TikTok; \textdollar190 billion\footnote{https://www.researchandmarkets.com/reports/5781298/social-media-market-report}), and interactive media (e.g., video games; \textdollar450 billion\footnote{https://www.statista.com/topics/868/video-games/}). While it is difficult to provide a concrete figure for the total annualized revenue of these industries collectively, one estimate forecasts that by 2029 the worldwide media and entertainment industry will grow to \textdollar3.5 trillion USD.\footnote{https://www.pwc.com/gx/en/issues/business-model-reinvention/outlook/insights-and-perspectives.html} 

Large AI companies are already beginning to pivot to developing AI for entertainment, despite marketing their products as general problem solvers. Recently, OpenAI reached an agreement to license Disney’s characters---from Mickey Mouse to Captain America to Yoda---for use in the Sora platform. A selection of these user-prompted videos will then be made available on the Disney+ streaming platform. Disney’s CEO, Robert Iger, noted in an official statement that ``Technological innovation has continually shaped the evolution of entertainment, bringing with it new ways to create and share great stories with the world.''\footnote{https://openai.com/index/disney-sora-agreement/} This is not just about using AI to quickly and cheaply create more movies based on Disney IP; the intention is to create entirely new ways of interactively engaging with characters and settings from Disney’s beloved fictional worlds. Beyond Sora, other recent releases from big AI companies are conspicuously geared towards generating entertaining content, such as Google’s Nano Banana and Meta’s Vibes.

As much as there may be an outcry to prevent AI from running rampant in creative industries, businesses in these industries seem eager to incorporate AI in production pipelines. For example, news outlets---which have been in the entertainment business since the advent of the 24 hour news cycle---are incorporating AI into their reporting practices. The \textit{BBC} uses AI for highly templated content, such as audio weather reports providing hyperlocal forecasts.\footnote{https://www.bbc.co.uk/aboutthebbc/reports/policies/what-were-doing-with-ai/} They are also piloting GenAI in relatively low-stakes settings, such as for generating ``bespoke audio bulletins'' offering match summaries for five English football clubs. The \textit{Financial Times} is piloting a chatbot where subscribers can query the newspaper’s archive of digitized content and receive a bespoke summary based on human reporting.\footnote{https://aboutus.ft.com/press\_release/financial-times-launches-first-generative-ai-tool} Though there are clear concerns about reliability, some aspects of AI are indeed well suited to these tasks: for example, synthesizing the gist of a text passage and adopting the authoritative, abstracted writing style of traditional reportage. 

GenAI will be used both to replace existing aspects of the entertainment production pipeline, as well as to expand it. AI represents a next major step in the steady creep of entertainment into daily life as it enlarges the space of possible formats entertainment can take---as with Sora-generated interactions with Disney characters or the \textit{BBC}’s hyperlocal weather reports. This shift will have a significant impact on society in everything from how news is made and consumed, to the influence that fictional characters can have on children and adolescents, to the amount and quality of attention-siphoning content available on our phones. Some reports have begun to suggest that AI’s alleged potential for economic disruption has been difficult to convert into actual productivity gains.\footnote{https://mlq.ai/media/quarterly\_decks/v0.1\_State\_of\_AI\_in\_Business\_2025\_Report.pdf}. By contrast, the adoption of AI as entertainment well underway. If companies find it difficult to get traction in selling AI tools for workplace productivity, more pressure will be put on them to generate revenue from entertainment. 

\section{Current AI evaluations look at the benefits of intelligence, harms of culture}

We argue that current AI evaluation techniques tend to examine the benefits of intelligence but the harms of culture. This represents a crucial gap in our approach to evaluating AI: the space in which we would articulate what we want out of AI in the cultural dimension, not just a description of what can go wrong. 

The standard procedures in AI evaluation were developed to measure the capabilities of systems engaging in well-defined tasks with clear metrics for success \cite{mehrotra2025understanding, becerra2025historical, russell1995modern}. AI outputs tend to be evaluated in much the same way as one would grade a student exam. There are right and wrong answers, and overall performance is judged by assigning a percentage score denoting how close the system got to some idealized response. This works well for many kinds of problems. For example, it works well when there is an unequivocal ground truth against which to compare an answer (e.g., whether or not a cat is pictured in a given image) or when there’s a well-defined value function denoting success (e.g., accrual of points in a video game). But this does not reflect the current landscape of use for GenAI systems. Mainstream approaches in AI evaluation are not well suited for outputs where there is no clear delineation between ``right'' and ``wrong'' answers, or when the criteria for success are ambiguous, contested, or highly context dependent \cite{kommers2025computational}. This is a crucial consideration in developing a positive vision of AI for entertainment. 

By comparison, social media platforms are widely understood to have had a detrimental societal effect on polarization \cite{aimeur2023fake}, mental health \cite{kross2021social}, and other important psychological or social factors \cite{fisher2022chaos}. One way to read this is as the downstream effect of an inadequate theory of what makes content worthwhile \cite{kommers2025meaning}. Social media algorithms prioritized content based on engagement. This kept users on the platforms for longer, providing more attentional surface area on which ads could be posted. While AI will likely not be subject to the same problem in the same form, if we do not supply a better theory of what makes content meaningful then a comparable set of unintended negative consequences is likely to result. We need a framework for articulating what makes an AI output actively meaningful in the cultural dimension---and not simply free from bias, prejudice, or the reinforcement of an inequitable power structure.

The arts and humanities are full of potential insights about what separates meaningful from mindless cultural content \cite{underwood2025impact, hemment2025doing, kommers2025computational}. For instance, one quality which is widely appreciated in the humanities but difficult to articulate within AI is friction \cite{shklovsky2015art, chen2024exploring, benford2012uncomfortable, garrett2025friction}. For technologists, friction is a negative attribute of a system. Good design consists in eliminating friction as the user progresses through an interface \cite{norman2013design}. But one manifestation in which this impulse goes unchecked is chatbot sycophancy---in which LLMs are fine-tuned to be maximally agreeable, responding with enthusiasm and even a degree of awe no matter what its human interlocutor says \cite{fanous2025syceval}. By contrast, in the humanities, friction is widely viewed as a crucial part of the meaning-making process \cite{shklovsky2015art}. Resistance is often what makes an experience memorable or transformative, for example when a novel presents a familiar theme or experience in a new light. We do not yet have a language for articulating---let alone quantitatively evaluating---how to calibrate for a quality like dialogic friction in GenAI models \cite{chen2024exploring, hemment2024experiential}.

A full survey of evaluation methods is beyond the scope of this paper. Instead, we take as a starting point a recent paper by Burden and colleagues \cite{burden2025paradigms}, who systematically evaluate different paradigms for AI evaluation. We focus on three of the paradigms they identify: benchmarks measuring task performance, such as the ability to solve a set of domain-specific problems; safety evaluations measuring systemic risk, such as a system's propensity to propagate misinformation; and construct-oriented tasks measuring system capabilities, such as psychological batteries for LLMs. We take these to be a representative set of some of the primary approaches (though by no means the only ones) by which we judge the performance of contemporary GenAI systems.

\begin{figure}[h]
  \centering
  \includegraphics[width=\linewidth]{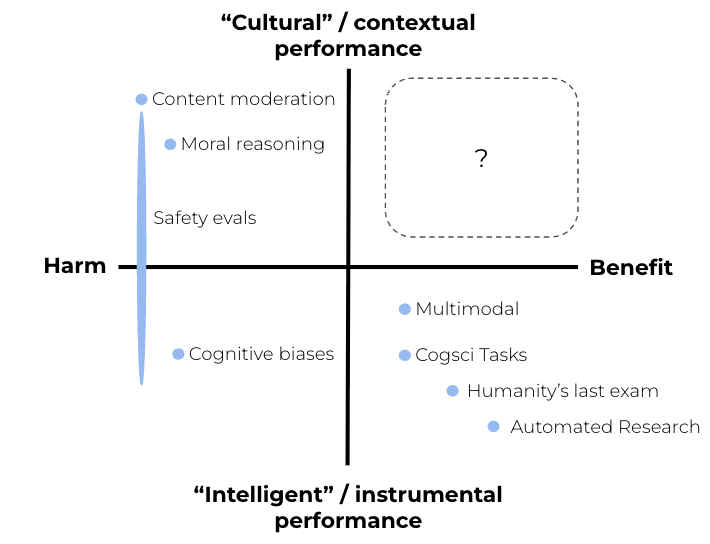}
  \caption{Contemporary evaluation protocols tend to focus on the benefits of intelligence and the harms of culture. Here we show a rough conceptual sketch organizing a subset of evaluations from a survey by Burden and colleagues (2025) \cite{burden2025paradigms} along two axes: harm vs benefit and cultural vs intelligent performance. We distinguish between intelligent performance as the ability to execute some sort of goal-oriented, instrumental process and cultural performance as the ability to abide by contextually relevant norms or accord with subjective human judgments. For example, multimodal test suites \cite{patraucean2023perception}, cogsci tasks \cite{coda2024cogbench}, Humanity's Last Exam \cite{phan2025humanity}, and benchmarks for automated AI researchers \cite{si2024can} evaluate the potential benefits of intelligence. By contrast, safety evaluations span both culture and intelligence but by definition focus exclusively on harms. This figure is not intended to serve as an authoritative delineation between culture and intelligence or to provide a comprehensive account of extant evaluation protocols. Rather, it meant to illustrate our claim that there is a relative paucity of AI evaluations that look at the benefits of culture.}
  \Description{A plot with two axes: the x dimension goes from harm to benefit, the y dimension goes from intelligent/instrumental performance to cultural/contextual performance. The intelligent-benefit quadrant is populated with example evaluations like multimodal evals, cogsci tasks, humanity's last exam, and protocols for automated research. The cultural-harm quadrant is populated with example evaluations like content moderation, safety evals, and moral reasoning. Safety evals also span into the intelligent-harm quadrant, along with cognitive baises. The cultural-benefit quadrant features an outlined rectangle with a question mark in the middle.}
  \label{fig:evals}
\end{figure}

Based on the evaluations surveyed by Burden and colleagues, there is clearly a rigorous standard for defining intelligence. Some of these take the form of benchmarks. For example, there are standardized tests for low-level cognitive functions in multimodal models \cite{patraucean2023perception}. These assess whether AI systems can go beyond static categorization or object detection to perform perceptual tasks like object tracking (e.g., did the ball move positions?), predictions about real-world physics (e.g., which configuration of blocks is more stable?), and relational reasoning (e.g., did the person put the teabag inside the cup or next to it?). Likewise, in the category of safety evaluations, there are robust tests for whether models can engage in strategic deception, using complex reasoning to deceive human users \cite{scheurer2024large}. In construct-oriented tasks, evaluations such as CogBench test the cognitive functions of LLMs using versions of the same experimental paradigms that would be used to test those of human undergraduates \cite{coda2024cogbench}. There are also tests of domain-specific expertise, such as Humanity’s Last Exam \cite{phan2025humanity}, as well as applications of intelligence for practical problem solving, such as automating parts of the scientific method \cite{si2024can, evans2025after}. Together, these represent a coherent, diverse, and increasingly robust set of standards for articulating a wide range of benefits and harms of intelligence.

There are also clear and rigorous standards for evaluating potential cultural harms. We define cultural performance as the ability to abide by contextually relevant norms or accord with subjective judgments made by humans. An example of the difference between evaluations of culture versus intelligence would be the assessment of moral reasoning \cite{franken2024procedural} versus cognitive biases \cite{koo2024benchmarking}. The study of moral reasoning is highly dependent on social context: different cultures have different standards of right and wrong. By definition, moral reasoning must be studied with at least some constraints on the social context in which the findings are meant to apply. By contrast, a cognitive bias such as confirmation bias or base rate neglect is not dependent on social context. It may manifest differently in different cultures, but the baseline judgment---whether someone seeks disconfirmatory evidence or ignores how base rates impact statistical inference---are based on third-party, non-social frames of reference. Assessments of both moral reasoning and cognitive biases test for a deviation from a standardized norm, rather than the achievement of uniquely good performance. Tests of moral reasoning typically are not meant to detect moral genius, as if on the look out for the next Confucius, Imhotep, Mother Theresa, or Dhrukpa Kunley. Rather they tend to place consensus-driven standards for appropriate behavior as the gold standard, then seek to measure violations in which an agent deviates from this baseline. In this sense, they might be better labeled as tasks of immoral reasoning. 

Safety evaluations often also have a cultural or contextual component \cite{ropers2024towards, meinke2024frontier}. Any test of potential misalignment necessarily poses the question: alignment to whom? Any answer must be based on a contextually-bound social judgment of value \cite{lazar2023ai, leibo2025societal, sorensen2024roadmap, yang2025alignment}. Likewise, content moderation also requires making contestable judgments about what kind of content should be allowed in a given context of discourse \cite{levi2025ai, akbulut2024century}. There are also batteries for assessing social bias or prejudice \cite{wan2023biasasker}. Together, these represent an emerging standard for identifying and mitigating potential cultural harms in GenAI systems.

In Figure~\ref{fig:evals}, we plot a subset of the evaluation methodologies surveyed by Burden and colleagues. Our plot reorganizes these evaluations into two dimensions: (1) whether they examine ``cultural'' or contextual performance versus ``intelligent'' or instrumental performance; and (2) whether they examine harms or benefits. The surveyed assessments cover a range of potential harms and benefits when it comes to intelligence. But for culture, they tend to focus exclusively on potential harms. This characterization of the landscape of AI evaluations is heuristic and by no means comprehensive. Any generalization about the current start of the art in AI evaluation is likely to be obsolete or incomplete before the ink on the publication is dry. And there are no doubt some interesting and exciting counterexamples \cite{shanahan2025xeno, underwood2025can, chen2024exploring, mire2025social, hamilton2025narrabench, lee2023evaluating}. However, our characterization points to an important trend in mainstream approaches to evaluating AI systems: we do not know what we want from AI in the cultural dimension.

\section{``Thick Entertainment'' as a vehicle for human meaning}

In this section, we offer an answer to what we might want out of AI in the cultural dimension. We call it ``thick entertainment.'' This perspective rests on the idea that the stories we consume, often via entertainment, play a central role in shaping how we interpret the meaning of our experiences \cite{kommers_sense-making_2025}. This is the social function that entertainment serves: a communal resource for meaning-making. It gives us narrative templates for how to make sense of our past experiences, just as it gives us exemplars for structuring future behavior and decisions. At least this is the function entertainment serves at its best. Some entertainment is purely brainrot---the equivalent of candy for the brain---while some entertainment is more nutritious. 

Our claim is that as AI is increasingly used for the purposes of entertainment, we need better ways of making such cultural and psychological meaning legible within the broader sociotechnical systems in which they are deployed \cite{kommers2025meaning}. As it stands, there is no scale way to measure ``meaningfulness'' at scale. This is one of the reasons why a theory of evaluation based on engagement prevailed for social media platforms at the expense of something deeper: it is a lot easier to measure. It is important to note that this does not equate to a claim that all entertainment must be Shakespearean in quality or that we are calling for removal of all works that fail to meet some elite standard of literary merit (e.g., runaway best-selling works of romantasy like Rebecca Yarros’s \textit{Fourth Wing}) from library shelves. On the contrary, it is possible for allegedly ``lowbrow'' culture to be a crucial source of meaning-making, even when it is AI-generated \cite{kommers2025slopmatters}. The difference here is on the margin. At the scale at which AI-generated entertainment will be deployed, it matters whether we have a way of discerning whether an underlying narrative or a symbolic image will challenge, or support, someone’s worldview in a constructive way \cite{hullman2023artificial}. 

People are willing to consume challenging entertainment. For example, one of the most commercially and critically successful TV series of 2025 was Stephen Graham’s \textit{Adolescence}. Across four episodes, the story follows the experiences of a British family as a 13-year-old boy is accused of murder. The show is challenging aesthetically (with each episode being filmed in a single take), and it is challenging thematically (dealing with hard questions about the role and limits of a parent’s ability to influence the direction of their child’s life). The show won five Emmy awards (including outstanding limited series); it was also one of the year’s most viewed shows on Netflix.\footnote{https://www.theguardian.com/tv-and-radio/2025/jun/03/adolescence-tv-series-netflix} To develop a positive theory of entertainment is to ask how AI-based systems can be designed and used in a rapidly evolving media ecosystem to support thick entertainment that is challenging and constructive, rather than incentivize work that merely provides fodder for the mindless haze of diversion without substance. 

Our proposed distinction between ``thick'' and ``thin'' entertainment derives from cultural anthropologist Clifford Geertz’s concept of thick description \cite{Geertz1973, kommers2025meaning, edelman2025full}. Geertz argues that cultural meaning can only be codified by providing an account of the social context in which a given symbol, action, ritual, or practice occurs. For example, a swastika has a very different meaning in the West, where it is associated with Nazi Germany, than in the East, where it is a symbol of prosperity and well-being in Hinduism, Jainism, and Buddhism. The framework of thick description is one of the methodological cornerstones of the humanities and qualitative social sciences \cite{savin2023qualitative}. It has previously been argued that this framework can help provide an alternative to ``thin'' metrics---such as views, shares, and comments---for evaluating the content produced and shared in large sociotechnical systems, such as social media, as well as emerging ones based on AI \cite{kommers2025meaning}.

Our position is not to advocate for a specific definition of thickness or standard for meaningfulness in entertainment. However, we can offer a sketch of potential ways of measuring thickness in entertaining content. These are not meant to be finalized recommendations, but rather to provide a few samples from the space of possibilities. One simple solution is just to avoid thinness: any attempt to use a graph with a line that goes up and to the right as unassailable evidence of progress is misguided. There are going to be multiple lines that matter. More often than not, progress in one dimension comes as a tradeoff to progress in another. It will probably be necessary to calibrate a range of dimensions that are relevant for entertainment. We already mentioned friction, but another example is ambiguity \cite{gaver2003ambiguity}. Thick entertainment does not reduce a theme or insight to a trivialized tagline; it leaves room for interpretation. At present, ambiguity is another quality that we typically seek to minimize in AI, rather than trying to find a sweet spot. 

Finally, a key aspect of how people interpret the meaning of their experience is the difference between judgments in the moment and judgments in retrospect \cite{kommers_sense-making_2025}---what Daniel Kahneman called the experiencing self versus remembering self \cite{kahneman2005living}. Engagement metrics such as views or likes reveal the preferences of the experiencing self but do not canvass the perspective of the remembering self. Developing measures that vary temporally could be a useful mechanism for assessing thickness in AI use. Overall, the big mistake here would be choosing to optimize AI systems for something just because it is what we know how to measure.

\section{Discussion}

AI is typically evaluated in terms of intelligence. While such a perspective on AI’s instrumental utility is valuable, it misses an important use of AI: entertainment. People are already using AI for entertainment. Myriad platforms, some with millions of users, employ AI for activities and tasks that have nothing to do with the kind of supercharged productivity for which the potential of AI is so often lauded. In this paper, we argue that we do not yet have a means of evaluating AS as entertainment. We contend that this will come to be seen as a dramatic oversight as AI upends digital entertainment ecosystems. In the long run, entertainment may even prove to be a more prevalent consumer use case for AI than workplace productivity.

The uptake of AI as entertainment is already underway. For example, emerging data suggests people’s motivation to use AI for entertainment is potentially larger and more durable than for use in goal-directed tasks \cite{chandra2025longitudinal}. We argue that this will offer a key business model for the large corporations building state-of-the-art AI systems. The potential disruption of entertainment-based industries may offer revenue streams of the size needed to justify the huge amount of money currently being invested in GenAI. Without a positive view of what it means to design AI for entertainment, the direction of this technology will be determined mainly by the bottom line of large corporations. 

There is a crucial gap in the way AI systems are evaluated. Mainstream approaches to evaluation tend to look at the benefits of intelligence and the harms of culture. Evaluations that look at socially-situated behaviors tend to take a strong position on what a system should not do (e.g., make prejudiced judgments or reinforce systemic bias) but typically do not offer a perspective on what we should want from AI in the cultural dimension \cite{lazar2023ai, weidinger2022taxonomy, sorensen2024roadmap}. While regulations and ethical guidelines are important, the pursuit of harm mitigation is a relatively weak posture to take in an effort to affect our society’s most powerful sociotechnical system.

We argue that we need a constructive vision of cultural AI---not just a theory of harm minimization, but a positive theory of what beneficial, nutritious entertainment might look like. There will be aspects of this that can leverage what AI is already good at, while also pointing towards areas of weakness where AI needs to be improved or supplemented. Under the right conditions, people are eager to consume challenging narratives en masse: stories with complex, nuanced thematic and aesthetic content. The problem is that we do not have a mechanism for making the meaningfulness of such content legible within AI-driven sociotechnical systems \cite{kommers2025meaning}. We contend that this positive vision of AI as entertainment should inspire more debates, discourse, and study in the field of AI. Generative AI is already, and will increasingly be, used for entertainment whether we like it or not. Academics and researchers should take an active role in shaping this application of AI technologies. If we don’t, we may come to regret it.

\subsection{Counterarguments}

In this section, we consider potential counterarguments to the position offered in this paper. Each of the following subsections begins with our attempt to briefly articulate a strong version of a potential counterargument. We then provide a response to this alternative perspective.

\subsubsection{Technologists don't dictate what gets made in Hollywood. There's no need to start just because people are using AI.} Creating and analyzing entertaining content is not the purview of science and engineering. It is the purview of artists, creatives, and cultural critics. They are simply separate jobs, and mission creep from engineering into this territory wouldn't solve anything---it would just muddy the waters. The most likely outcome of this proposal is to crowd out creative people we want to support rather than replace.

Our response: We would emphasize that our position is not that we should make AI maximally entertaining. It is also not that one kind of entertainment matters more or is better than another. Instead, our position is that entertainment is a use case which the field of AI is largely unprepared to evaluate. We don't suggest that AI startups should all suddenly pivot to a business model in which they act as subcontractors for Hollywood production or AI slop content farms. Our point is that the seemingly small decisions about how AI systems are designed will have a major impact not just in how people use these tools at work, but about the kind of narratives they hold about the world. We saw how algorithmic manipulation of this harder-to-pin-down notion affected society with social media. If technologists say this isn't their job, then we're failing to learn from our previous mistakes.

\subsubsection{It isn't possible for an entertainment-based business model to be morally or ethically superior. It all depends on how it is implemented, governed, and regulated.}

This claim about ethically aligned business models is illusory. Considerations like ethics, morality, and human flourishing are by nature illegible to any given business model, because a business model succeeds or fails solely on the basis of whether it makes money. The reason academics don't offer suggestions about business models is because there is no way to create a business model that reliably or intentionally elicits the societal benefits we want. We have to look to other avenues, like regulation.

Our response: We accept this as a legitimate position within this space of arguments. Our larger point is that discussions of AI business models should be a part of academic discourse, because over time this will be the consideration which most impacts the kind of technology that corporations produce. However, such a binary view of the effects of one business model over another doesn't seem like the most useful model of reality. For example, one could imagine a coffee company whose value proposition is to provide the cheapest coffee possible, independent of quality or ethical sourcing. By contrast, there are many companies whose explicit value proposition is to deliver a high quality product while treating farmers fairly: they plan to make money specifically by offering a product that benefits as many people as possible in its supply chain. Would the world be a better place if we gave up on this distinction because in the end all coffee companies want to make money? 

\subsubsection{We don't want AI to be good at making meaningful entertainment. Let's leave that to the humans.}

Yes, AI will be used to make entertainment. Yes, there is potential to make money here. But the problem is that this will come at the expense of the humans who already make entertainment (i.e., artists). This is a bad outcome; we should be doing everything we can to stave it off, not rush toward it. 

Our response: We agree with the motivation behind this concern and agree it's an important area to scrutinize closely. In many creative industries, there are major concerns that entry-level jobs will be or are already being eliminated. In one potential scenario, this will lead to the degradation of the foundational skills of the people with the most influence in arts and entertainment. In another scenario, this will lead to fewer people---or fewer kinds of people---who will can pursue a career in this field. In another scenario, AI will continually creep up the arts and entertainment production line, being used for more and more aspects of creative output. Perhaps all of these could happen in conjuction; we don't want to see any of them come to fruition. However, we think it's important to draw a distinction between what we would like to see happen within the arts and entertainment industry, and what the powerful forces of technology are realistically going to do to it. Our account is offered the latter spirit. There is too much money at stake and entertainment is too alluring of a market for AI companies not to pursue it. If we really care about protecting artists and creatives, the best way to do that is to have an open and rigorous discussion about AI as entertainment and what we want that to look like---not to pretend like it isn't going to happen.

\subsubsection{Part of the issue with misinformation in social media was the selective presentation of facts in echo chambers. Aren't you setting us up to create a comparable problem with isolated islands of hyper-personalized meaning-making?}

This suggestion that entertainment should be made as meaningful as possible sounds great but in reality would come with a lot of issues. Drawing on the example of social media, it could be argued that social media platforms did in fact become ecosystems for maximally meaningful content. Take the example of conspiracy theories like QAnon. Social media created the conditions for the ``meaning'' of otherwise banal, unrelated content to be weaved into webs of significance. This seems a lot like what is being suggested with ``thick entertainment'' (or at least something that could potentially come as an unintended consequence). Beyond that, just as social media has created hyper-personalized information spheres, where different factions of society are presented with entirely different facts, it seems like an effect of this proposal will be to do this in the dimension not of belief and knowledge but of meaning-making and storytelling. This doesn't seem like it would end well.

Our response: This is a major challenge to the basic premise of our argument for thick entertainment. One potential outcome of creating meaningful entertainment that can be tailored to individual perspectives and circumstances is that it could diminish the common reservoir of shared stories across society. This is a bedrock of social organization. It is important not just that people endorse certain stories as meaningful to themselves personally, but that other people share their endorsement. The motivation for our argument is that the composition of this cultural fabric matters, and we need to study how AI is going to affect it. With that in mind, it is nonetheless possible that any attempt to increase the potential positive benefits of AI as entertainment will come with concomitant negative impacts as well. However, on balance, we argue the right move here is to create systems that are sensitive to these kind of impacts, rather than ignore them. This is the thrust of ``thick'' entertainment: not to say that thickness consists in a specific set of variables, but that we need to look at more than just engagement.

\subsubsection{In fact, we do know what we want from cultural AI: pluralism.}

Any attempt to say what is a ``better'' form of entertainment will necessarily come at the expense of another equally legitimate form. Trying to police culture in this way is a bad idea. Instead, the best thing we can do is give people the opportunity to make their own choice. Once you start prioritizing one kind of content or another, you're going to marginalize some voices while giving a platform to others. There is no good way to do this in a top-down manner. Even if there was, technologists probably shouldn't be the people response for implementing it.

Our response: No, pluralism is not the only thing we want from cultural AI. It is true that policing culture is usually a bad idea. But the right response is not simply to resign from any perspective on evaluating the quality of content. If a completely free market of culture was possible to achieve, maybe that would indeed be the ideal state. But that's far from the reality of our situation. Content is going to be prioritized in one way or another: by pursuit of profit, by existing power dynamics, by ancillary effects of algorithmic design. Pluralism is important; it is a crucial consideration in designing systems for the kind of thick entertainment we are proposing. But to say that pluralism is the only thing we can aim for is to abdicate responsibility over the impacts a given system will have. Ultimately it would be a mistake not to have difficult conversations about what kind of entertaining content we think will be beneficial or detrimental for society.

\section{Generative AI Usage Statement}

The main body of this manuscript was draft without the help of generative AI. However, we used it to draft a first version of the abstract, which we then edited ourselves.


\bibliographystyle{ACM-Reference-Format}
\bibliography{main}

\end{document}